\documentclass[journal,10pt,twocolumn]{IEEEtran}
\usepackage{graphicx}
\usepackage{amsmath,amsfonts}
\usepackage{amssymb}
\usepackage{newtxtext,newtxmath}
\usepackage[caption=false,font=normalsize,labelfont=sf,textfont=sf]{subfig}
\usepackage{cite}
\usepackage{makecell}
\usepackage{algorithm}
\usepackage{algorithmic}
\usepackage{array}
\usepackage{amsmath,bm,amsfonts}
\usepackage{url,color}
\usepackage{epsfig,multirow,algorithm,setspace,float}
\usepackage{booktabs}
\usepackage{setspace}
\usepackage{hyperref}
\usepackage{orcidlink}
\hypersetup{
    colorlinks=true,
    linkcolor=blue,
    filecolor=magenta,
    urlcolor=cyan,
    citecolor=blue
}
\usepackage{float}

\newcommand\myfigpath{.}

\makeatletter
\renewcommand{\maketag@@@}[1]{\hbox{\m@th\normalsize\normalfont#1}}
\makeatother

\usepackage{titlesec}
\usepackage{caption}
\usepackage{enumitem}

\titlespacing*{\section}{0pt}{1.2ex plus 0.5ex minus 0.2ex}{0.8ex plus 0.2ex}
\titlespacing*{\subsection}{0pt}{1.0ex plus 0.4ex minus 0.2ex}{0.6ex plus 0.2ex}
\titlespacing*{\subsubsection}{0pt}{0.8ex plus 0.3ex minus 0.2ex}{0.5ex plus 0.2ex}

\setlist[itemize]{topsep=2pt, itemsep=1pt, parsep=0pt, partopsep=0pt, leftmargin=1.2em}
\setlist[enumerate]{topsep=2pt, itemsep=1pt, parsep=0pt, partopsep=0pt, leftmargin=1.4em}

\renewcommand{\arraystretch}{0.95}

\begin{document}	
	
\title{Physics-Aligned Spectral Mamba: Decoupling Semantics and Dynamics for Few-Shot Hyperspectral Target Detection}

\author{Luqi~Gong\textsuperscript{\orcidlink{0009-0000-8744-8630}},~\IEEEmembership{Student Member,~IEEE,}
Qixin~Xie\textsuperscript{\orcidlink{0009-0002-4457-966X}},
Yue~Chen\textsuperscript{\orcidlink{0009-0008-4975-4854}},
Ziqiang~Chen\textsuperscript{\orcidlink{0009-0004-5138-9893}},
Fanda~Fan\textsuperscript{\orcidlink{0000-0002-5214-0959}},~\IEEEmembership{Member,~IEEE,}
Shuai~Zhao\textsuperscript{\orcidlink{0000-0002-5217-004X}},~\IEEEmembership{Member,~IEEE,}
and Chao~Li\textsuperscript{\orcidlink{0000-0002-5343-1862}},~\IEEEmembership{Member,~IEEE}
\thanks{Corresponding authors: Chao Li, Shuai Zhao.}
\thanks{Luqi Gong is with the Research Center for Space Computing System, Zhejiang Lab, Hangzhou 311121, China, and also with the State Key Laboratory of Networking and Switching Technology, Beijing University of Posts and Telecommunications, Beijing 100876, China (e-mail: luqi@zhejianglab.org).}
\thanks{Qixin Xie, Yue Chen, Ziqiang Chen, and Chao Li are with the Research Center for Space Computing System, Zhejiang Lab, Hangzhou 311121, China (e-mail: keshawn.hsieh@zhejianglab.org; chenyue@zhejianglab.org; chenzq@zhejianglab.org; lichao@zhejianglab.org).}
\thanks{Fanda Fan is with the Institute of Computing Technology, Chinese Academy of Sciences, Beijing 100190, China (e-mail: fanfanda@ict.ac.cn).}
\thanks{Shuai Zhao is with the State Key Laboratory of Networking and Switching Technology, Beijing University of Posts and Telecommunications, Beijing 100876, China (e-mail: zhaoshuaiby@bupt.edu.cn).}
}

\maketitle

\begin{abstract}
Meta-learning facilitates few-shot hyperspectral target detection (HTD), but adapting deep backbones remains challenging. Full-parameter fine-tuning is inefficient and prone to overfitting, and existing methods largely ignore the frequency-domain structure and spectral band continuity of hyperspectral data, limiting spectral adaptation and cross-domain generalization.
To address these challenges, we propose SpecMamba, a parameter-efficient and frequency-aware framework that decouples stable semantic representation from agile spectral adaptation. Specifically, we introduce a Discrete Cosine Transform Mamba Adapter (DCTMA) on top of frozen Transformer representations. By projecting spectral features into the frequency domain via DCT and leveraging Mamba’s linear-complexity state-space recursion, DCTMA explicitly captures global spectral dependencies and band continuity while avoiding the redundancy of full fine-tuning. Furthermore, to address prototype drift caused by limited sample sizes, we design a Prior-Guided Tri-Encoder (PGTE) that allows laboratory spectral priors to guide the optimization of the learnable adapter without disrupting the stable semantic feature space. Finally, a Self-Supervised Pseudo-Label Mapping (SSPLM) strategy is developed for test-time adaptation, enabling efficient decision boundary refinement through uncertainty-aware sampling and dual-path consistency constraints. Extensive experiments on multiple public datasets demonstrate that SpecMamba consistently outperforms state-of-the-art methods in detection accuracy and cross-domain generalization.

\end{abstract}
	
	
	\IEEEpeerreviewmaketitle

	\section{Introduction}
	
	\begin{figure}[!t]
		\centering
		\includegraphics[width=1\linewidth]{\myfigpath/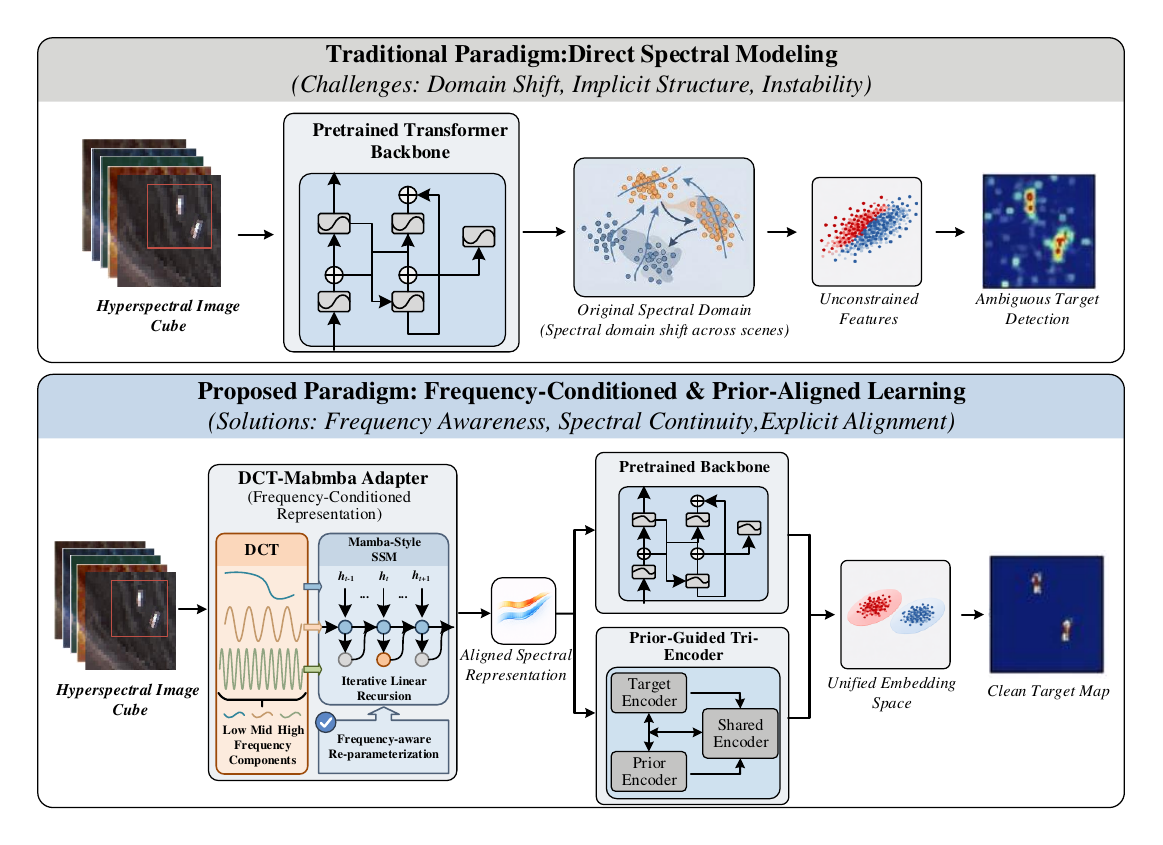}
		\caption{Motivation of the proposed SpecMamba. Conventional direct spectral modeling often leads to boundary ambiguity, while the proposed frequency-conditioned modeling reorganizes HSI representations in the frequency domain to effectively separate targets from background.}
		\label{moti}
	\end{figure}
	
\IEEEPARstart{H}{yperspectral} imaging (HSI) provides rich spectral measurements that enable fine-grained discrimination of materials beyond what is achievable with conventional RGB imagery~\cite{ZHU2023205}. By leveraging the distinctive spectral signatures of target materials, hyperspectral target detection (HTD) aims to identify specific objects of interest embedded in complex and cluttered backgrounds~\cite{10349920}. Such capability has made HTD a critical tool in a wide range of applications, including environmental monitoring, urban analysis, and security-related scenarios~\cite{fang2025guided}.
Despite its potential, practical HTD is often conducted under highly constrained conditions. In many real-world deployments, only a few reference spectral signatures of the target are available, while the observed scenes may vary significantly across sensing platforms, acquisition conditions, and background compositions~\cite{8113122}. As a result, HTD is inherently a few-shot and cross-domain problem, where effective detection must be achieved with extremely limited supervision and strong distribution shifts.

Recent deep learning methods for HTD typically rely on powerful semantic backbones to learn discriminative representations from data~\cite{Wang_2022_meta}. In this paradigm, convolutional or transformer-based networks are usually pretrained on source scenes or auxiliary datasets and then adapted to new target scenes through fine-tuning~\cite{Dong_2024_06}. To alleviate the shortage of annotations, data augmentation~\cite{YANG2020107464}, metric learning~\cite{6658949}, meta-learning, and transfer learning~\cite{asg,10210405} have been widely explored. However, most methods adapt the entire network or a large subset of parameters using only a few target samples~\cite{Wang_2022_meta}, which entangles stable semantic representation learning with task-specific adaptation. Under limited supervision, this often leads to overfitting and poor generalization, revealing a conflict between representation stability and adaptation flexibility in few-shot HTD~\cite{Zhao_2024_03}.

Existing HTD frameworks also remain limited in spectral modeling. As shown in Fig.~\ref{moti}, most deep learning methods~\cite{jiao2023tt, he2025spectral, 10205510} treat hyperspectral data as stacked channels and model them through channel-wise convolutions, spectral attention, or tokenized transformer sequences, often together with spatial interactions~\cite{jiao2023tt}. Although effective, these strategies usually encode spectral bands implicitly and do not explicitly exploit the continuity of hyperspectral signatures~\cite{JIAO2023109016}. In practice, discriminative cues depend not only on individual bands, but also on global spectral shape and long-range wavelength dependency. Neglecting such frequency-domain structure and band continuity can produce noise-sensitive and domain-unstable representations under scarce supervision~\cite{10205510}. This limitation motivates explicit spectral modeling for robust few-shot HTD.

These observations suggest that an effective few-shot HTD framework~\cite{10402099,10980209} should satisfy three requirements: parameter-efficient adaptation to preserve stable semantics under limited supervision; explicit spectral modeling to capture the frequency-domain structure and continuity of hyperspectral signatures~\cite{11082338,10225559,10313066}; and linear-complexity modeling for high-dimensional spectral sequences. From this perspective, frequency-domain adaptation offers a compact and physically meaningful way to describe global spectral characteristics, while sequential modeling enables efficient long-range dependency learning. As shown in Fig.~\ref{moti}, unlike direct spectral modeling that often suffers from feature entanglement and boundary ambiguity, the proposed frequency-conditioned paradigm better separates target signatures from background clutter.

Motivated by these challenges, we propose \textbf{SpecMamba}, a parameter-efficient and frequency-aware framework for few-shot HTD. SpecMamba decouples stable semantic representation from task-specific spectral adaptation by using a pretrained Transformer encoder as a frozen semantic backbone and performing adaptation through a lightweight spectral adapter. To explicitly model spectral structure and band continuity, we further introduce a \textbf{Discrete Cosine Transform Mamba Adapter (DCTMA)}, which combines DCT-based frequency representation with Mamba-based state-space modeling for linear-complexity long-range dependency learning. In addition, laboratory spectral priors are incorporated to regularize adapter optimization and improve cross-domain alignment without disturbing the frozen semantic space. During inference, an uncertainty-aware self-supervised refinement strategy is adopted to adjust the decision boundary to the target scene. Extensive experiments show that SpecMamba achieves superior accuracy and robustness in diverse few-shot settings.

    The main contributions of this article are summarized as follows:
    \begin{enumerate}
    	\item We propose the \textbf{Discrete Cosine Transform Mamba Adapter (DCTMA)}, a parameter-efficient spectral adaptation module that combines DCT-based frequency decomposition with linear-complexity state-space modeling to capture global spectral dependencies and band continuity with low computational cost.
    	
    	\item We introduce a \textbf{Prior-Guided Tri-Encoder (PGTE)} with a bifurcated alignment mechanism to separate stable semantic representation from task-specific spectral adaptation, allowing laboratory spectral priors to regularize the learnable adapter and reduce semantic drift in few-shot cross-domain settings.
    	
    	\item We develop an \textbf{uncertainty-aware Self-Supervised Pseudo-Label Mapping (SSPLM)} strategy for test-time adaptation, which selectively updates the lightweight adapter and detection head under a parameter-freezing scheme to improve robustness in cluttered target scenes.
    	
    	\item Extensive experiments on public HTD benchmarks show that SpecMamba achieves superior detection accuracy and cross-domain generalization, verifying the effectiveness of the proposed frequency-aware and parameter-efficient design.
    \end{enumerate}
    
    The remainder of this article is organized as follows. Section~\ref{sec:related_work} reviews related work. Section~\ref{sec:intro3} presents the proposed SpecMamba framework. Section~\ref{sec:intro4} reports experimental and ablation results on real hyperspectral datasets. Section~\ref{sec:intro5} concludes this article.

\section{Related Work}
\label{sec:related_work}

\subsection{Conventional Statistical and Subspace Detectors}
Classical HTD methods are mainly based on statistical hypothesis testing and subspace decomposition. Early spectral similarity measures, including spectral angle mapper (SAM) \cite{KRUSE1993} and spectral information divergence (SID) \cite{Chein2000}, provide fundamental criteria for target discrimination. To suppress background interference, adaptive coherence estimator (ACE) \cite{Kraut1999} and orthogonal subspace projection (OSP) \cite{Harsanyi1994, Chang2005OSP} exploit background covariance or orthogonal subspaces for detection.

Another widely used paradigm is constrained energy minimization (CEM) \cite{Du2003, Chang2000CEM}, which has been extended to hierarchical \cite{Zou2015Hierarchical}, ensemble-based \cite{Zhao2019Ensemble}, and kernelized forms \cite{Kwon2006Kernel}. More recent studies further improve robustness to spectral variability through collaborative representation \cite{CSHTD}, sparse discrimination \cite{LU2018}, and hybrid low-rank decomposition \cite{9781337}.

\subsection{Deep Learning-based HTD}
Deep learning has significantly advanced HTD \cite{10919111,10980219,10976430} through hierarchical feature extraction. Early methods mainly relied on supervised convolutional networks \cite{li2017td} and autoencoders \cite{zhang2020htd}, while later studies introduced interpretable architectures, such as HTD-IRN \cite{Shen2023HTDIRN} and HTD-Net \cite{HTDNET2020}. Transformer-based models, including TSTTD \cite{jiao2023tt} and STTD \cite{ran2022st}, further improve discrimination by modeling global spectral dependencies.

To alleviate the demand for labeled samples, weakly supervised and self-supervised strategies have been widely explored, such as HTDFormer \cite{Li2023HTDFormer}, spectral-level contrastive learning \cite{wang2023ss}, and generative masking methods. More recent works incorporate deformable transformers \cite{Fang2025Deformable}, deep-growing networks with manifold constraints \cite{Shi2025DeepGrowing}, unsupervised momentum contrastive learning \cite{Wang2024Unsupervised,11236413}, and adaptive sample generation \cite{asg} to enhance discriminative capability under limited supervision.

\subsection{Few-shot and Transfer Learning in HTD}
Few-shot learning and transfer learning are essential for HTD, since detectors must generalize from label-rich source domains to data-scarce target scenes under notable distribution shifts. To alleviate such gaps, prior studies have explored sensor-independent learning \cite{shi2021si}, siamese meta-learning \cite{Wang2022Meta}, regularized tensor representation \cite{10066274}, and background learning with target suppression constraints \cite{Xie2021Background} to improve generalization in unseen environments.

Nevertheless, most transfer-oriented methods \cite{10811951} still treat deep backbones as black-box feature extractors. Without explicitly considering the physical consistency of spectral signatures across heterogeneous sensors, they are prone to catastrophic forgetting and overfitting to local background clutter during target-domain adaptation \cite{LIU2026112030}. This limitation highlights the need to decouple stable semantic representation from task-specific spectral adaptation, so as to better balance stability and plasticity.

\subsection{State Space Models for Hyperspectral Analysis}
State Space Models (SSMs), especially Mamba \cite{Gu2023Mamba}, provide an efficient alternative to Transformers for modeling long-range dependencies. Unlike self-attention with quadratic complexity, Mamba employs structured state-space transitions \cite{Gu2021Efficiently} for linear-time sequence modeling, making it well suited to high-dimensional hyperspectral data.

For hyperspectral analysis, Mamba can effectively capture global spectral dependencies and band continuity while suppressing redundant information \cite{jiang2025sedgm, he20243dss}. Its selective scanning mechanism further supports scalable spectral-spatial feature modeling \cite{wang2024spectral}, offering a promising solution under limited training data. Recent efforts have also combined state-space modeling with meta-learning \cite{chakrabarty2023meta, xiao2025cascaded} to improve feature initialization and adaptation robustness in cross-domain settings.

\begin{figure*}[!t]
	\centering
	\includegraphics[width=0.9\linewidth]{\myfigpath/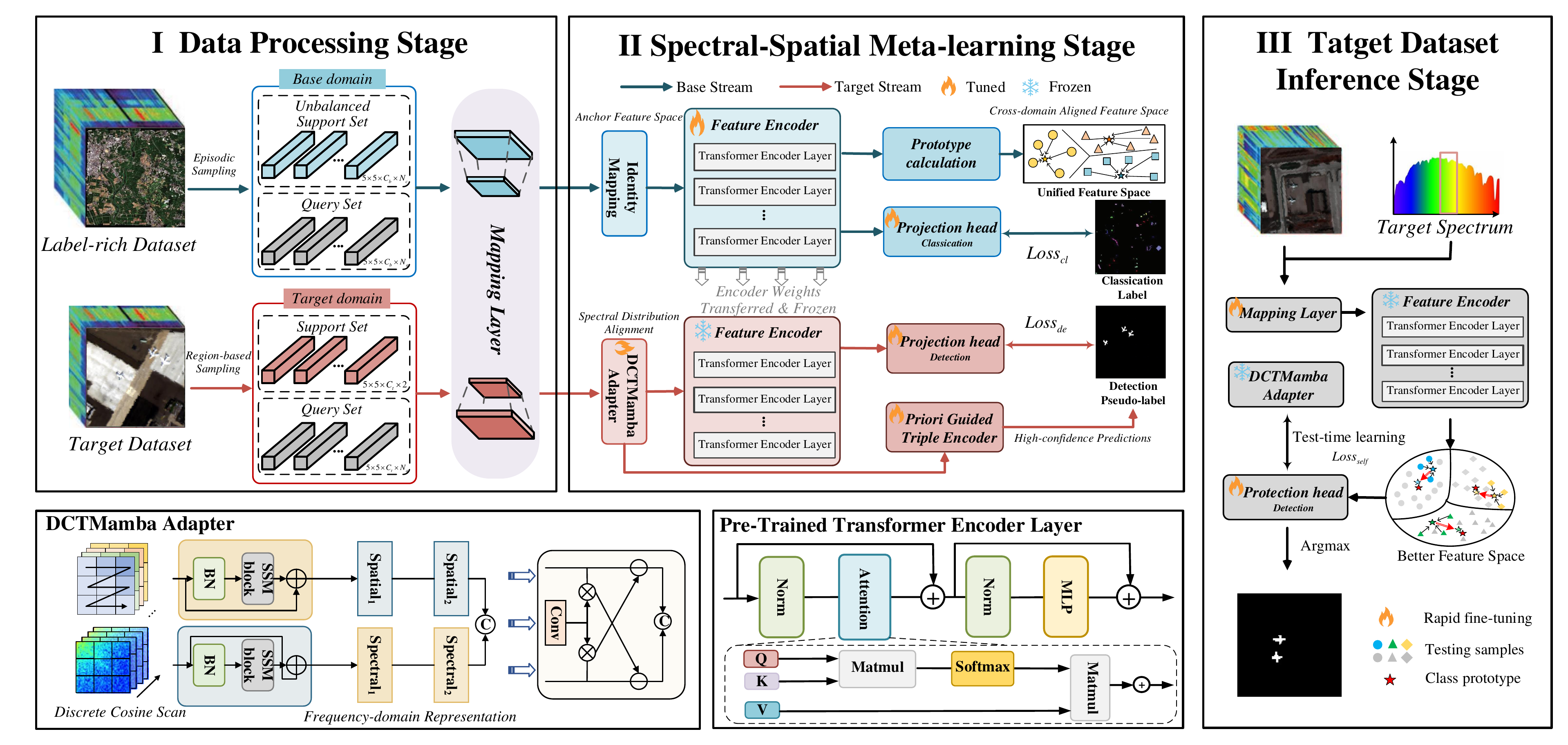}
	\caption{Overview of the proposed SpecMamba framework. It consists of three stages: data processing, dual-stream spectral-spatial meta-learning with DCTMA and PGTE, and target-domain inference with SSPLM.}
	\label{fig:framework}
\end{figure*}

\section{METHODOLOGY}\label{sec:intro3}

\subsection{Overall Framework}
\label{subsec:framework}

SpecMamba addresses few-shot HTD under cross-domain spectral shifts. As shown in Fig.~\ref{fig:framework}, it contains two phases: spectral-spatial meta-learning and uncertainty-aware inference. In the first phase, a Transformer encoder $\mathcal{F}_{\Theta}$ is trained on the source domain to learn a universal prototype space. For the target domain, $\Theta$ is transferred and frozen, while a lightweight DCTMA is introduced for adaptation. Its output is split into a perception path for frozen-encoder embedding extraction and an alignment path for PGTE-based physical alignment. In the second phase, SSPLM performs test-time adaptation on the target scene by updating only the adapter and detection head using high-confidence pseudo-labels with weighted pseudo-supervision and self-supervised consistency.

\subsection{Problem Formulation}
\label{subsec:data_prep}

A hyperspectral image is denoted by $\mathbf{X}\in\mathbb{R}^{H\times W\times B}$, where $H\times W$ and $B$ represent the spatial size and spectral bands, respectively. For each pixel $(i,j)$, we extract a local patch $\mathbf{x}_{i,j}\in\mathbb{R}^{s\times s\times B}$ as the basic input unit. Meta-learning is performed on the label-rich base domain $\mathcal{D}_{base}$ using $N$-way $K$-shot episodes. Each task $\mathcal{T}$ contains a support set $\mathcal{S}$ and a query set $\mathcal{Q}$:
\begin{equation}
	\mathcal{S}=\{(\mathbf{x}_k^s,y_k^s)\}_{k=1}^{N\times K},\quad
	\mathcal{Q}=\{(\mathbf{x}_m^q,y_m^q)\}_{m=1}^{M_q},
\end{equation}
where $K$ is the number of support samples per class. The objective is to learn encoder parameters $\Theta$ such that query samples are classified by similarity to support prototypes, yielding a class-agnostic metric space. In the target domain $\mathcal{D}_{target}$, we are given a target image $\mathbf{X}_t$ and a small set of clean target spectra $\mathbf{T}_{prior}=\{\mathbf{t}_k\}_{k=1}^{K}$, without background labels. Unlike the multi-class setting in $\mathcal{D}_{base}$, the target task is binary detection, and binary pseudo-episodes are constructed via the rebalanced pseudo-labeling strategy in Section~\ref{subsec:ssplm}.

\subsection{Frequency-Aware Spectral-Spatial Modeling via DCTMA}
\label{subsec:DCTMA}

Extracting robust features for few-shot HTD faces two contradictory hurdles: (1) \textbf{Spectral Redundancy}: HSI bands are highly correlated, where discriminative absorption features are often submerged in noise; (2) \textbf{Computational Cost}: Capturing global spatial context usually requires heavy computations. To resolve this, we propose the \textit{Discrete Cosine Transform Mamba Adapter} (DCTMA), an architecture that unifies frequency-domain spectral analysis with linear-complexity spatial state-space modeling.

\subsubsection{Frequency-Domain Spectral Decomposition}
Directly processing raw spectral curves often leads to overfitting due to high inter-band correlation and noise. We propose to re-parameterize the spectral profile into the frequency domain, explicitly separating the background continuum from fine-grained material signatures.
Given a spectral-spatial patch $\mathbf{x}_i \in \mathbb{R}^{s \times s \times B}$ centered at the $i$-th pixel, we first flatten its spatial dimensions to obtain a spectral-dominant sequence $\bar{\mathbf{x}}_i \in \mathbb{R}^{B \times s^2}$. A 1-D Discrete Cosine Transform (DCT) is applied along the spectral dimension (rows):
\begin{equation}
\mathbf{f}_i = \mathcal{D}(\bar{\mathbf{x}}_i) = \text{DCT}(\bar{\mathbf{x}}_i), \quad \mathbf{f}_i \in \mathbb{R}^{B \times s^2}.
\end{equation}
In the frequency domain, energy compaction properties indicate that smooth background variations are concentrated in low-frequency coefficients, whereas sharp absorption features that are critical for target identification reside in mid- and high-frequency bands. To exploit this structural prior, we partition the frequency spectrum indices $\mathcal{I} = \{1, \dots, B\}$ into three disjoint sets via split ratios $0 < \rho_L < \rho_M < 1$:
\begin{align}
\mathcal{I}_L &= \{1, \dots, \lfloor \rho_L B \rfloor\}, \nonumber \\
\mathcal{I}_M &= \{\lfloor \rho_L B \rfloor + 1, \dots, \lfloor \rho_M B \rfloor\}, \\
\mathcal{I}_H &= \{\lfloor \rho_M B \rfloor + 1, \dots, B\}. \nonumber
\end{align}
We then apply hard masking matrices $\mathbf{M}^{(g)} = \text{diag}(m_1, \dots, m_B)$ for each group $g \in \{L, M, H\}$, where $m_k=1$ if $k \in \mathcal{I}_g$ and 0 otherwise. This yields three frequency-specific components $\mathbf{f}_i^{(g)} = \mathbf{M}^{(g)} \mathbf{f}_i$.

To capture the distinct semantic information of each band group, we employ a lightweight encoder $\psi_g(\cdot)$ to generate a compact descriptor $\mathbf{z}_i^{(g)} \in \mathbb{R}^{d}$. This encoding process essentially maps the high-dimensional frequency coefficients to a latent metric space, formulated as:
\begin{equation}
\mathbf{z}_i^{(g)} = \psi_g(\mathbf{f}_i^{(g)}) = \text{Pool}\left( \sigma(\mathbf{W}_g \mathbf{f}_i^{(g)}) \right),
\end{equation}
where $\mathbf{W}_g$ is a learnable projection matrix, $\sigma(\cdot)$ denotes the activation function, and $\text{Pool}(\cdot)$ represents the global average pooling operation that aggregates information across the spatial tokens. 

Furthermore, since the importance of each frequency range varies across materials, we introduce an \textit{Adaptive Spectral Gating} mechanism. An attention vector $\mathbf{w}_{att}$ computes the importance scores via Softmax:
\begin{equation}
\alpha_i^{(g)} = \frac{\exp(\mathbf{w}_{att}^\top \mathbf{z}_i^{(g)})}{\sum_{h \in \{L, M, H\}} \exp(\mathbf{w}_{att}^\top \mathbf{z}_i^{(h)})}.
\end{equation}

The final frequency-aware spectral embedding is the dynamic weighted sum:
\begin{equation}
\mathbf{E}^{(spec)}_i = \sum_{g \in \{L, M, H\}} \alpha_i^{(g)} \mathbf{z}_i^{(g)}.
\end{equation}

\begin{figure*}[!t]
	\centering
	\includegraphics[width=0.85\linewidth]{\myfigpath/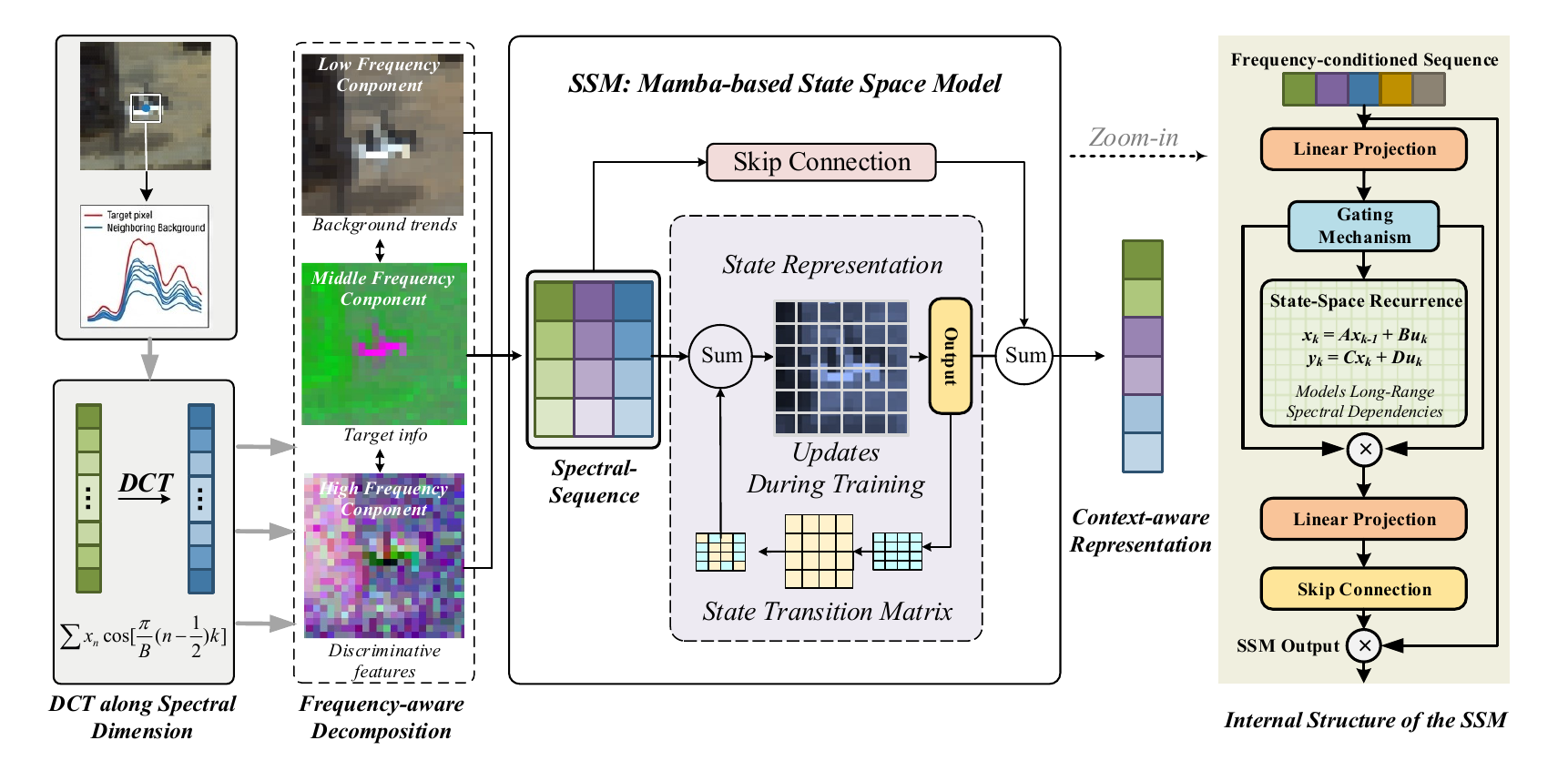}
	\caption{Detailed architecture of the selective SSM in the DCTMA module. It combines DCT-based frequency decomposition with Mamba-based state-space modeling to capture long-range spectral dependencies and suppress background clutter.}
	\label{fig:moudle}
\end{figure*}

\subsubsection{Linear-Complexity Spatial Modeling via Selective SSM}
While the spectral branch handles material identification, the spatial branch is responsible for structural consistency and clutter suppression. Traditional Transformers scale quadratically with sequence length. We instead leverage the Mamba architecture, which relies on Selective State Space Models (SSM) to achieve global receptive fields with linear complexity $O(N)$, as illustrated in Fig. \ref{fig:moudle}.

We reshape the input patch into a flattened spatial token sequence $\hat{\mathbf{x}}_i \in \mathbb{R}^{s^2 \times B}$. The core of Mamba is a continuous-time system that maps a 1-D input function $x(t)$ to $y(t)$ through a hidden state $h(t)$:
\begin{equation}
h'(t) = \mathbf{A}h(t) + \mathbf{B}x(t), \quad y(t) = \mathbf{C}h(t).
\end{equation}
Here, $\mathbf{A}$ is the evolution matrix and $\mathbf{B}, \mathbf{C}$ are projection parameters. To implement this in deep learning, we employ the Zero-Order Hold (ZOH) discretization rule with a dynamic timescale parameter $\Delta$. The discretized system parameters are derived as follows:
\begin{align}
\overline{\mathbf{A}} &= \exp(\Delta \mathbf{A}), \\
\overline{\mathbf{B}} &= (\Delta \mathbf{A})^{-1}(\exp(\Delta \mathbf{A}) - \mathbf{I}) \cdot \Delta \mathbf{B}.
\end{align}
The recurrence relation for the spatial token sequence at step $t$ becomes:
\begin{equation}
\mathbf{h}_t = \overline{\mathbf{A}} \mathbf{h}_{t-1} + \overline{\mathbf{B}} \hat{\mathbf{x}}_{i,t}, \quad \mathbf{y}_t = \mathbf{C} \mathbf{h}_t.
\end{equation}
Crucially, the timescale $\Delta$ is input-dependent, allowing the model to selectively focus on relevant spatial tokens while ignoring irrelevant clutter. By scanning the spatial sequence through this mechanism, we obtain a global context-aware representation $\mathbf{E}^{(spa)}_i = \text{Mamba}(\hat{\mathbf{x}}_i)$.

\subsubsection{Cross-Gated Modality Fusion}
Spectral features $\mathbf{E}^{(spec)}_i$ and spatial features $\mathbf{E}^{(spa)}_i$ offer complementary views. To ensure coherent integration, we propose a \textit{Cross-Gated Interaction} module where each modality modulates the other.
Let $\tilde{\mathbf{E}}^{(spec)}_i$ and $\tilde{\mathbf{E}}^{(spa)}_i$ denote the projected features broadcasting to compatible dimensions. The refined representations are computed via reciprocal modulation:
\begin{align}
\mathbf{H}^{(spec)}_i &= \tilde{\mathbf{E}}^{(spec)}_i \odot \sigma\left(\mathbf{W}_{spa \to spec} \tilde{\mathbf{E}}^{(spa)}_i\right), \\
\mathbf{H}^{(spa)}_i &= \tilde{\mathbf{E}}^{(spa)}_i \odot \sigma\left(\mathbf{W}_{spec \to spa} \tilde{\mathbf{E}}^{(spec)}_i\right),
\end{align}
where $\odot$ is the element-wise product and $\sigma$ is the Sigmoid function. This mechanism enables the spatial context to suppress spectral features in regions with high noise (e.g., boundaries).

Finally, we define the DCTMA output as the fused adapter feature $\mathbf{h}^{ada}_i$:
\begin{equation}
\label{eq:DCTMA_output}
\mathbf{h}^{ada}_i = \phi_{fuse}\left( [\mathbf{H}^{(spec)}_i, \mathbf{H}^{(spa)}_i] \right).
\end{equation}

This embedding $\mathbf{h}^{ada}_i$ serves as the robust spectral-spatial foundation for the entire framework. As illustrated in the system architecture, the data flow bifurcates at this point: $\mathbf{h}^{ada}_i$ is simultaneously fed into the \textbf{Prior-Guided Tri-Encoder (PGTE)} for physically consistent metric learning (Section \ref{subsec:pgte}) and the \textbf{Feature Encoder} for semantic detection tasks (Section \ref{subsec:ssplm}), establishing our proposed dual-path design.

\subsection{Physically Consistent Metric Learning via PGTE}
\label{subsec:pgte}

In standard few-shot detection, class prototypes are constructed solely by averaging the support set embeddings. However, in real-world HTD, support samples are often few and contaminated by noise, leading to \textit{prototype drift}. To rectify this, we propose the \textit{Prior-Guided Tri-Encoder} (PGTE), which introduces a laboratory-measured clean spectrum as a physical anchor to guide the adaptation of the DCTMA.

For a target category $c$, let $\mathbf{t}_{prior}$ denote its clean physical spectral signature (retrieved from a spectral library $\mathbf{T}_{prior}$). As shown in the system architecture, the PGTE operates through a tri-branch design:
\begin{itemize}
    \item \textbf{Prior Encoder ($\mathcal{E}_{p}$):} A multi-layer perceptron (MLP) that maps the physical signature $\mathbf{t}_{prior}$ into a stable anchor embedding $\mathbf{e}_{prior} = \mathcal{E}_{p}(\mathbf{t}_{prior})$.
    \item \textbf{Alignment Encoder ($\mathcal{E}_{a}$):} A learnable projection head that takes the bifurcated DCTMA output $\mathbf{h}^{ada}_i$ (from the Alignment Path defined in Sec. \ref{subsec:DCTMA}) as input. This branch aims to map the adapted features to the same metric space as the prior.
    \item \textbf{Frozen Backbone ($\mathcal{F}_{frozen}$):} Represents the main semantic branch used for final inference, extracting $\mathbf{e}_i$ from the Perception Path.
\end{itemize}

\subsubsection{Prior-Driven Prototype Rectification}
To ensure the learned DCTMA features respect physical reality, we impose a \textit{Physical Consistency Constraint} ($\mathcal{L}_{phy}$). This loss minimizes the distance between the adapted support features and the physical anchor:
\begin{equation}
\mathcal{L}_{phy} = \frac{1}{K} \sum_{k=1}^{K} \left\| \mathcal{E}_{a}(\mathbf{h}^{ada}_k) - \mathbf{e}_{prior} \right\|_2^2.
\end{equation}
Crucially, since $\mathcal{E}_{a}$ connects directly to the DCTMA output $\mathbf{h}^{ada}_k$, this loss allows gradients to bypass the frozen backbone and directly update the DCTMA parameters $\Theta_{DCTMA}$, forcing the adapter to align the raw data distribution with the physical prior.

Simultaneously, we construct a \textbf{Rectified Prototype} $\mathbf{p}_c$ for inference. Instead of relying solely on noisy support samples, we fuse the data-driven prototype with the physical anchor using a smooth balancing coefficient $\lambda$:
\begin{equation}
\mathbf{p}_c = \lambda \cdot \frac{1}{K}\sum_{k=1}^K \mathcal{F}_{frozen}(\mathbf{h}^{ada}_k) + (1-\lambda) \cdot \mathbf{e}_{prior}.
\end{equation}
This ensures the prototype remains robust even when support samples are heavily contaminated.

\subsection{Dual-Task Optimization Objective}
\label{subsec:optimization}

To learn a feature space that is both semantically discriminative (from Base Domain) and task-adaptive (for Target Domain), we employ a dual-task optimization strategy comprising two dedicated heads.

\subsubsection{Classification Loss ($Loss_{cl}$)}
On the \textbf{Base Stream}, the model learns general spectral semantics. We attach a Classification Head (a linear layer $\mathbf{W}_{cls}$) to the output of the trainable backbone $\mathbf{e}_i$. The objective is to maximize the separability among $N$ base categories via Cross-Entropy Loss:
\begin{equation}
Loss_{cl} = - \sum_{i \in \mathcal{Q}_{base}} \log \frac{\exp(\mathbf{w}_{y_i}^\top \mathbf{e}_i)}{\sum_{j=1}^{N} \exp(\mathbf{w}_j^\top \mathbf{e}_i)}.
\end{equation}
This loss prevents feature collapse and ensures the backbone $\mathcal{F}_{\Theta}$ learns a discriminative manifold.

\subsubsection{Detection Loss ($Loss_{de}$)}
On the \textbf{Target Stream}, the goal shifts to binary discrimination (Target vs. Background). We attach a Detection Head (a binary classifier $\mathbf{w}_{det}$) to the output of the \textit{frozen} backbone. Since ground-truth labels are unavailable, we utilize high-confidence pseudo-labels $\hat{y}_i \in \{0, 1\}$ generated by the rectified prototypes. The loss is formulated as Binary Cross-Entropy (BCE):
\begin{equation}
Loss_{de} = - \sum_{i \in \mathcal{Q}_{target}} \left[ \hat{y}_i \log(P_i) + (1 - \hat{y}_i) \log(1 - P_i) \right],
\end{equation}
where $P_i = \sigma(\mathbf{w}_{det}^\top \mathbf{e}_i)$ is the predicted target probability. This objective fine-tunes the DCTMA and Detection Head to suppress background clutter in the specific target scene.

\subsubsection{Total Objective}
The framework is trained in an end-to-end meta-learning manner. The total objective is a weighted sum of the semantic, detection, and physical constraints:
\begin{equation}
\mathcal{L}_{total} = Loss_{cl} + \beta Loss_{de} + \gamma \mathcal{L}_{phy},
\end{equation}
where $\beta$ and $\gamma$ are hyperparameters that balance the trade-off between semantic generalization, task specificity, and physical consistency.

\subsection{Uncertainty-Aware Adaptation via SSPLM}
\label{subsec:ssplm}

Although the meta-trained SpecMamba learns transferable features, direct inference on the target domain $\mathcal{D}_{target}$ is still affected by domain shifts caused by atmospheric variation and sensor discrepancy. To alleviate this without ground-truth annotations, we propose \textit{Self-Supervised Pseudo-Label Mapping} (SSPLM), which performs rapid test-time adaptation (TTA) by refining pseudo-labels and updating lightweight adapter modules.

\subsubsection{High-Confidence Pseudo-Label Generation}
Given the target hyperspectral image $\mathbf{X}_t$ and the rectified prototype $\mathbf{p}_c$ obtained by PGTE, we first compute a coarse similarity map $\mathbf{S} \in \mathbb{R}^{H \times W}$. For pixel $i$, the similarity score is defined by cosine similarity:
\begin{equation}
	s_i = \frac{(\mathbf{e}_i)^\top \mathbf{p}_c}{\| \mathbf{e}_i \| \| \mathbf{p}_c \|},
\end{equation}
where $\mathbf{e}_i$ is the embedding extracted by the frozen backbone.

To reduce error accumulation from noisy predictions, we adopt a \textit{Region-based Uncertainty Sampling} strategy with thresholds $\tau_{neg}$ and $\tau_{pos}$. Pixels with $s_i > \tau_{pos}$ are assigned to the positive pseudo-set $\Omega_{pos}$, while those with $s_i < \tau_{neg}$ form the negative pseudo-set $\Omega_{neg}$. Pixels within the uncertainty interval are discarded to avoid unstable updates. The pseudo-label $\hat{y}_i$ is set to 1 for $\Omega_{pos}$ and 0 for $\Omega_{neg}$.

\subsubsection{Spectral-Spatial Hybrid Augmentation}
To regularize learning on limited pseudo-labeled samples, we construct augmented views using a hybrid transformation $\mathcal{T}(\cdot)$ that includes:
\begin{itemize}
	\item \textbf{Spatial Transformation}: random rotation ($k \cdot 90^\circ$) and flipping;
	\item \textbf{Spectral Perturbation}: additive Gaussian noise $\epsilon \sim \mathcal{N}(0, \delta^2)$.
\end{itemize}
For each selected sample $\mathbf{x}_i$, an augmented view is generated as $\tilde{\mathbf{x}}_i = \mathcal{T}(\mathbf{x}_i)$, encouraging invariance to task-irrelevant perturbations.

\subsubsection{Dual-Constraint Adaptation Objective}
During inference, we adopt a parameter-efficient fine-tuning strategy by freezing the Feature Encoder and updating only the DCTMA parameters $\Theta_{DCTMA}$ and the Detection Head $\mathbf{w}_{det}$. This allows adaptation to the target scene while preserving the meta-learned representation. The optimization is guided by two objectives.

\paragraph{Weighted Pseudo-Supervision ($\mathcal{L}_{wbce}$)}
To address the severe class imbalance between background and target pixels, we use a weighted binary cross-entropy loss:
\begin{equation}
	\mathcal{L}_{wbce} = - \sum_{i \in \Omega} \omega_i \left[ \hat{y}_i \log P(\mathbf{x}_i) + (1 - \hat{y}_i) \log (1 - P(\mathbf{x}_i)) \right],
\end{equation}
where $\Omega = \Omega_{pos} \cup \Omega_{neg}$, $P(\mathbf{x}_i)$ denotes the predicted probability, and $\omega_i$ is inversely proportional to the class size.

\paragraph{Self-Supervised Consistency ($\mathcal{L}_{self}$)}
To exploit unlabeled information, we further impose consistency between the predictions of the original sample $\mathbf{x}_i$ and its augmented view $\tilde{\mathbf{x}}_i$:
\begin{equation}
	\mathcal{L}_{self} = \sum_{i \in \Omega} \left\| P(\mathbf{x}_i) - P(\tilde{\mathbf{x}}_i) \right\|_2^2.
\end{equation}
This regularization stabilizes the decision boundary under different perturbations.

The final TTA objective over the tunable parameters $\{\Theta_{DCTMA}, \mathbf{w}_{det}\}$ is
\begin{equation}
	\min_{\Theta_{DCTMA}, \mathbf{w}_{det}} \mathcal{J} = \mathcal{L}_{wbce} + \eta \mathcal{L}_{self}.
\end{equation}
After rapid fine-tuning (e.g., 50 iterations), the adapted model produces the final detection map for $\mathbf{X}_t$.

\section{EXPERIMENTAL RESULTS AND ANALYSIS}\label{sec:intro4}

\subsection{Experimental Setup}
\label{subsec:setup}

\subsubsection{Datasets and Data Partitioning}
To evaluate the generalization and adaptation ability of SpecMamba, we conduct experiments under a source-to-target transfer setting.

\begin{figure}[!htbp]
	\centering
	\includegraphics[width=0.7\linewidth]{\myfigpath/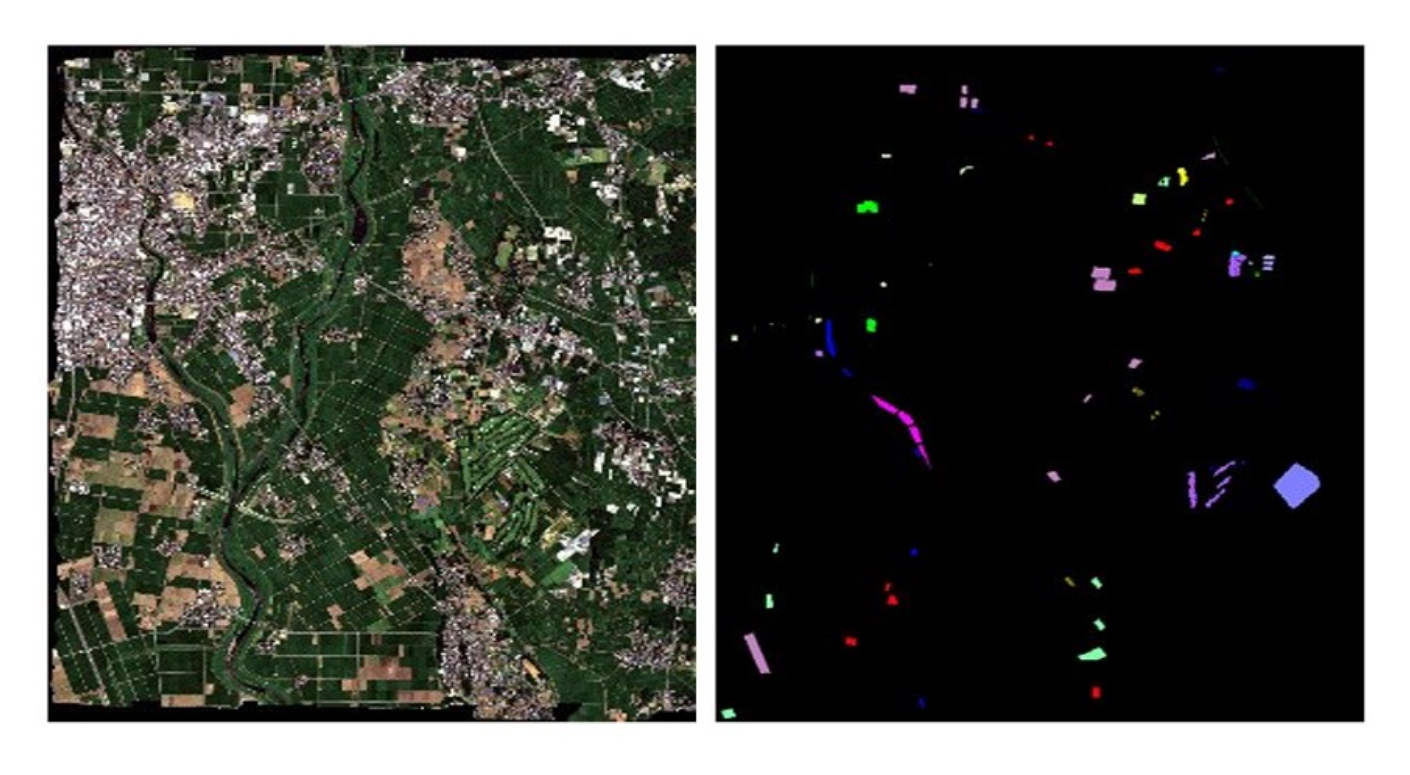}
	\caption{Pseudo-color image and Ground truth of the Source data (Chikusei).}
	\label{fig:source_data}
\end{figure}

\textbf{Source Domain ($\mathcal{D}_{base}$):} We use the \textbf{Chikusei} dataset, a large-scale aerial hyperspectral image with 128 spectral bands. To balance category distribution for meta-learning, classes with more than 5,000 samples (classes 4, 7, 8, 9, and 14) are excluded. The remaining 14 land-cover categories are used to construct episodic training tasks. The pseudo-color image and corresponding ground-truth map are shown in Fig.~\ref{fig:source_data}.

\begin{figure}[!htbp]
	\centering
	\includegraphics[width=0.7\linewidth]{\myfigpath/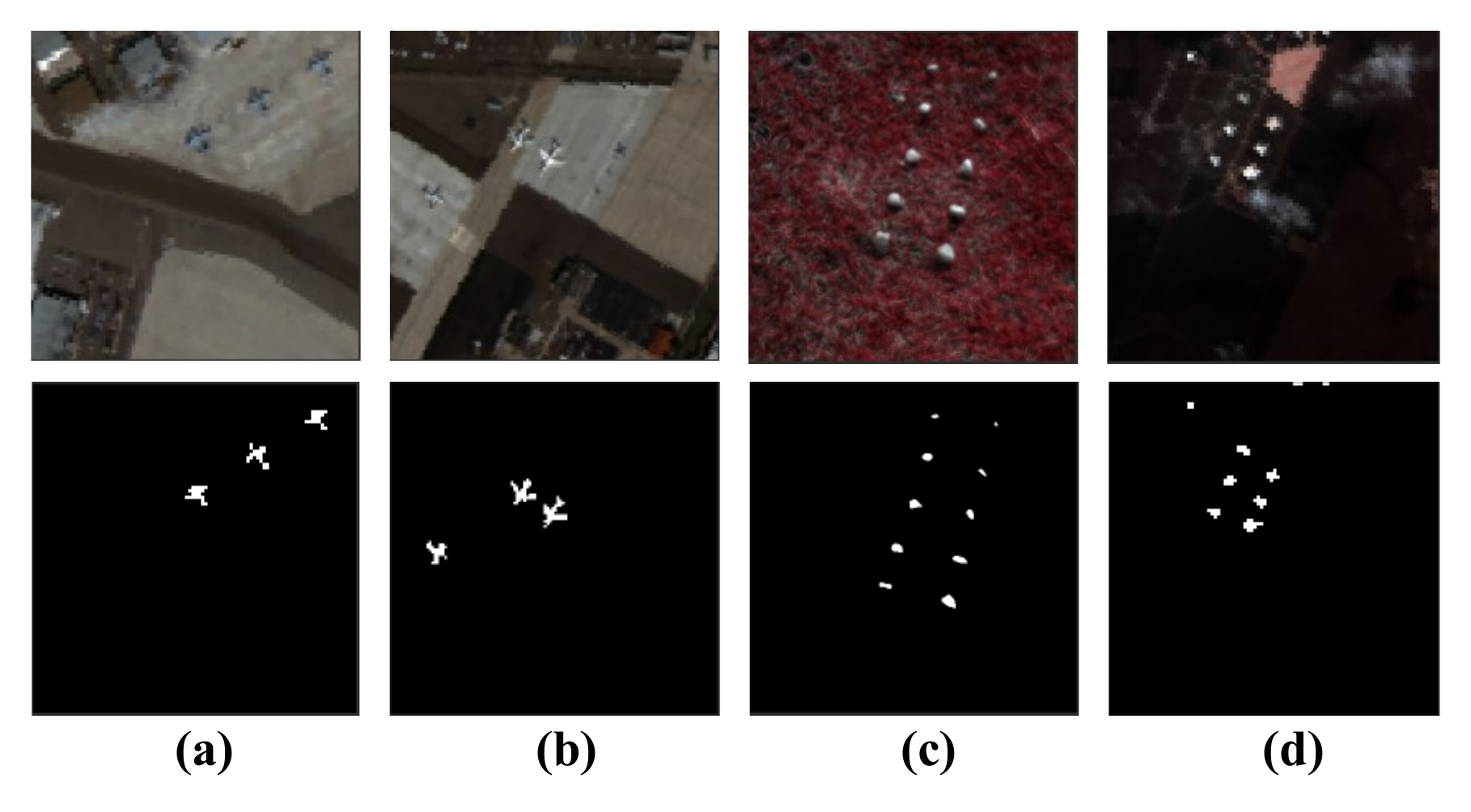}
	\caption{Dataset image scenes and ground truth. (a) San Diego I. (b) San Diego I. (c) Cir. (d) Urban.}
	\label{fig:target_data}
\end{figure}

\textbf{Target Domain ($\mathcal{D}_{target}$):} For few-shot HTD, we adopt four public benchmarks: \textbf{Sandiego I}, \textbf{Sandiego II}, \textbf{Cri}, and \textbf{abu-urban-1}. These datasets cover diverse sensing conditions and contain challenging sub-pixel targets in complex backgrounds. Their pseudo-color images and target annotations are shown in Fig.~\ref{fig:target_data}.

\begin{table*}[!htbp]
	\centering
	\caption{Quantitative comparison of different methods on four HSI datasets}
	\label{tab:quantitative}
	\footnotesize
	\setlength{\tabcolsep}{3pt}
	\renewcommand{\arraystretch}{1.12}
	
	\begin{tabular}{ccccccccccc}
		\toprule
		HSI Data Set & Metric
		& CEM~\cite{Du2003} & ACE~\cite{Kraut1999} & MLSN~\cite{wang2022ml} & BLTSC~\cite{xie2020bl} & HTD-IRN~\cite{shen2023ht} & UEML~\cite{LIU2026112030} & SelfMTL~\cite{SelfMTL} & HTD-Mamba~\cite{HTDMamba} & Ours \\
		\midrule
		
		\multirow{5}{*}{San Diego I}
		& AUC$(P_F,P_D)\,\uparrow$
		& 0.99751 & 0.99455 & 0.98844 & 0.99325 & 0.99613 & 0.99597 & 0.97869 & \underline{0.99861} & \textbf{0.99927} \\
		& AUC$(\tau,P_D)\,\uparrow$
		& 0.88664 & \underline{0.98047} & 0.89927 & 0.83205 & 0.65967 & 0.84379 & 0.17207 & 0.95748 & \textbf{0.98220} \\
		& AUC$(\tau,P_F)\,\downarrow$
		& 0.10539 & 0.07026 & 0.53271 & 0.23976 & \underline{0.01859} & 0.10251 & \textbf{0.00455} & 0.14622 & 0.16227 \\
		& AUC$_{\text{OA}}\,\uparrow$
		& 1.77877 & 1.27244 & 1.35500 & 1.58554 & 1.63721 & 1.73725 & 1.14622 & \underline{1.80986} & \textbf{1.81919} \\
		& AUC$_{\text{SNPR}}\,\uparrow$
		& 8.41339 & 11.39553 & 1.68811 & 3.47033 & \underline{35.48498} & 8.23123 & \textbf{37.86353} & 6.54818 & 6.05276 \\
		\midrule

		\multirow{5}{*}{San Diego II}
		& AUC$(P_F,P_D)\,\uparrow$
		& 0.91535 & 0.94235 & 0.89938 & 0.97314 & 0.94725 & 0.86952 & 0.90251 & \underline{0.97585} & \textbf{0.98915} \\
		& AUC$(\tau,P_D)\,\uparrow$
		& 0.48156 & \underline{0.96457} & 0.72291 & 0.26850 & 0.43449 & 0.47790 & 0.09462 & 0.51787 & \textbf{0.97598} \\
		& AUC$(\tau,P_F)\,\downarrow$
		& 0.24390 & 0.04389 & 0.53119 & \textbf{0.00225} & 0.07999 & 0.09076 & \underline{0.00253} & 0.09572 & 0.09592 \\
		& AUC$_{\text{OA}}\,\uparrow$
		& 1.15301 & 1.06304 & 1.09110 & 1.23938 & 1.30175 & 1.25666 & 0.99460 & \underline{1.39800} & \textbf{1.86921} \\
		& AUC$_{\text{SNPR}}\,\uparrow$
		& 1.97443 & 1.14302 & 1.36092 & \textbf{119.21777} & 5.43172 & 5.26571 & \underline{37.47907} & 5.41028 & 10.17494 \\
		\midrule
		
		\multirow{5}{*}{Cri}
		& AUC$(P_F,P_D)\,\uparrow$
		& 0.99987 & 0.99883 & 0.71869 & 0.99904 & 0.99979 & \underline{0.99990} & 0.99989 & 0.99947 & \textbf{0.99994} \\
		& AUC$(\tau,P_D)\,\uparrow$
		& 0.62624 & \underline{0.91116} & 0.74874 & 0.43837 & 0.72743 & 0.83151 & 0.25272 & 0.85104 & \textbf{0.93107} \\
		& AUC$(\tau,P_F)\,\downarrow$
		& 0.17643 & 0.61365 & 0.60596 & \underline{0.00581} & 0.20354 & 0.06067 & \textbf{0.00289} & 0.10719 & 0.10743 \\
		& AUC$_{\text{OA}}\,\uparrow$
		& 1.44968 & 1.45634 & 0.86147 & 1.43161 & 1.52368 & \underline{1.77074} & 1.25232 & 1.74332 & \textbf{1.82358} \\
		& AUC$_{\text{SNPR}}\,\uparrow$
		& 3.54956 & 1.48482 & 1.23562 & \underline{75.45506} & 3.57397 & 13.70510 & \textbf{885.12353} & 7.93929 & 8.66653 \\
		\midrule
		
		\multirow{5}{*}{Urban-1}
		& AUC$(P_F,P_D)\,\uparrow$
		& 0.52790 & 0.99168 & 0.97826 & 0.95326 & 0.98283 & \underline{0.99250} & 0.98569 & 0.98169 & \textbf{0.99942} \\
		& AUC$(\tau,P_D)\,\uparrow$
		& 0.49166 & \textbf{0.97965} & 0.80460 & 0.93151 & 0.37311 & 0.89683 & 0.35522 & \underline{0.97296} & 0.46699 \\
		& AUC$(\tau,P_F)\,\downarrow$
		& 0.46146 & 0.54405 & 0.17913 & 0.15514 & 0.03337 & 0.09487 & \underline{0.00894} & 0.16355 & \textbf{0.00101} \\
		& AUC$_{\text{OA}}\,\uparrow$
		& 0.55810 & 1.42728 & 1.60372 & \underline{1.72963} & 1.32257 & \textbf{1.79446} & 1.33196 & 1.79110 & 1.46541 \\
		& AUC$_{\text{SNPR}}\,\uparrow$
		& 1.06545 & 1.80067 & 4.49169 & 6.00433 & 11.18171 & 19.45329 & \underline{39.73696} & 5.94902 & \textbf{464.73765} \\
		\bottomrule
	\end{tabular}
\end{table*}

\begin{figure*}[!t]
	\centering
	\includegraphics[width=0.8\linewidth]{\myfigpath/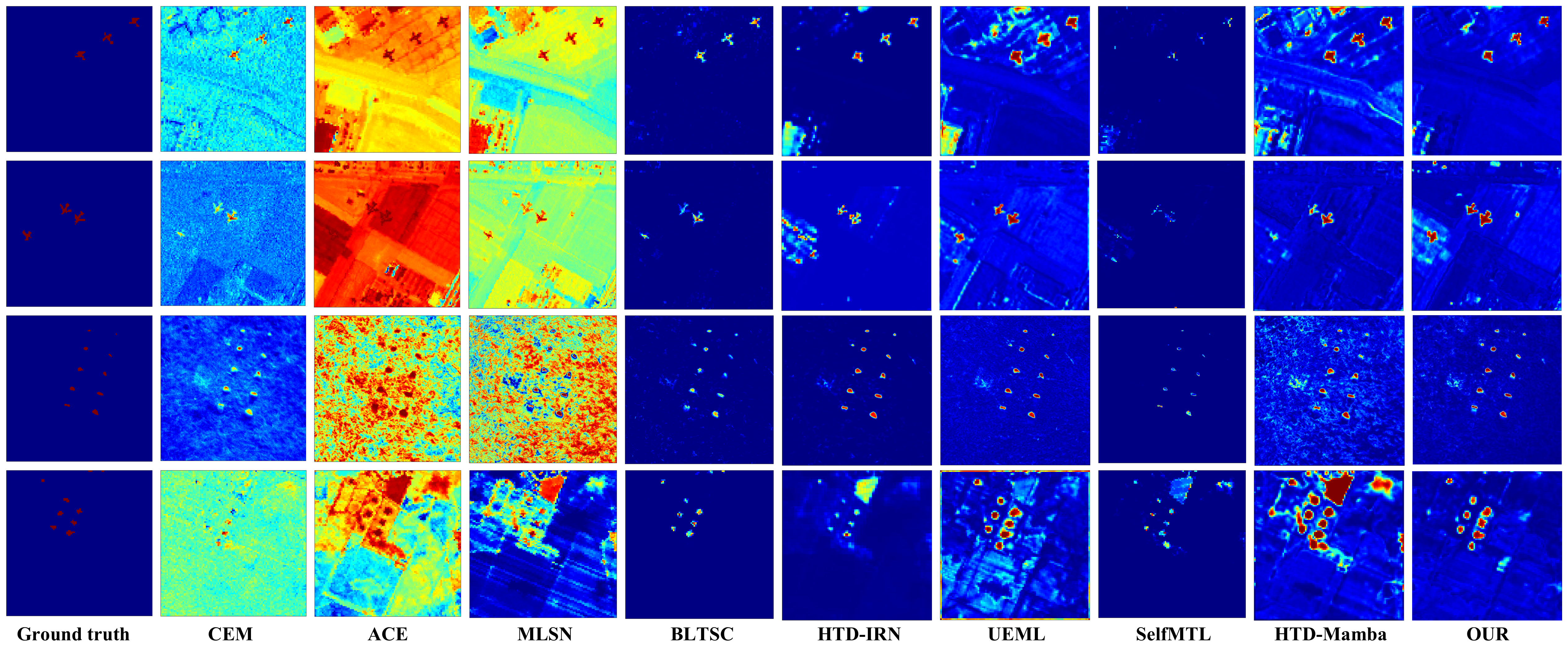}
	\caption{Experimental datasets and visualized detection maps of competing methods.}
	\label{fig:visual_maps}
\end{figure*}

\begin{figure*}[!t]
	\centering
	\includegraphics[width=0.8\linewidth]{\myfigpath/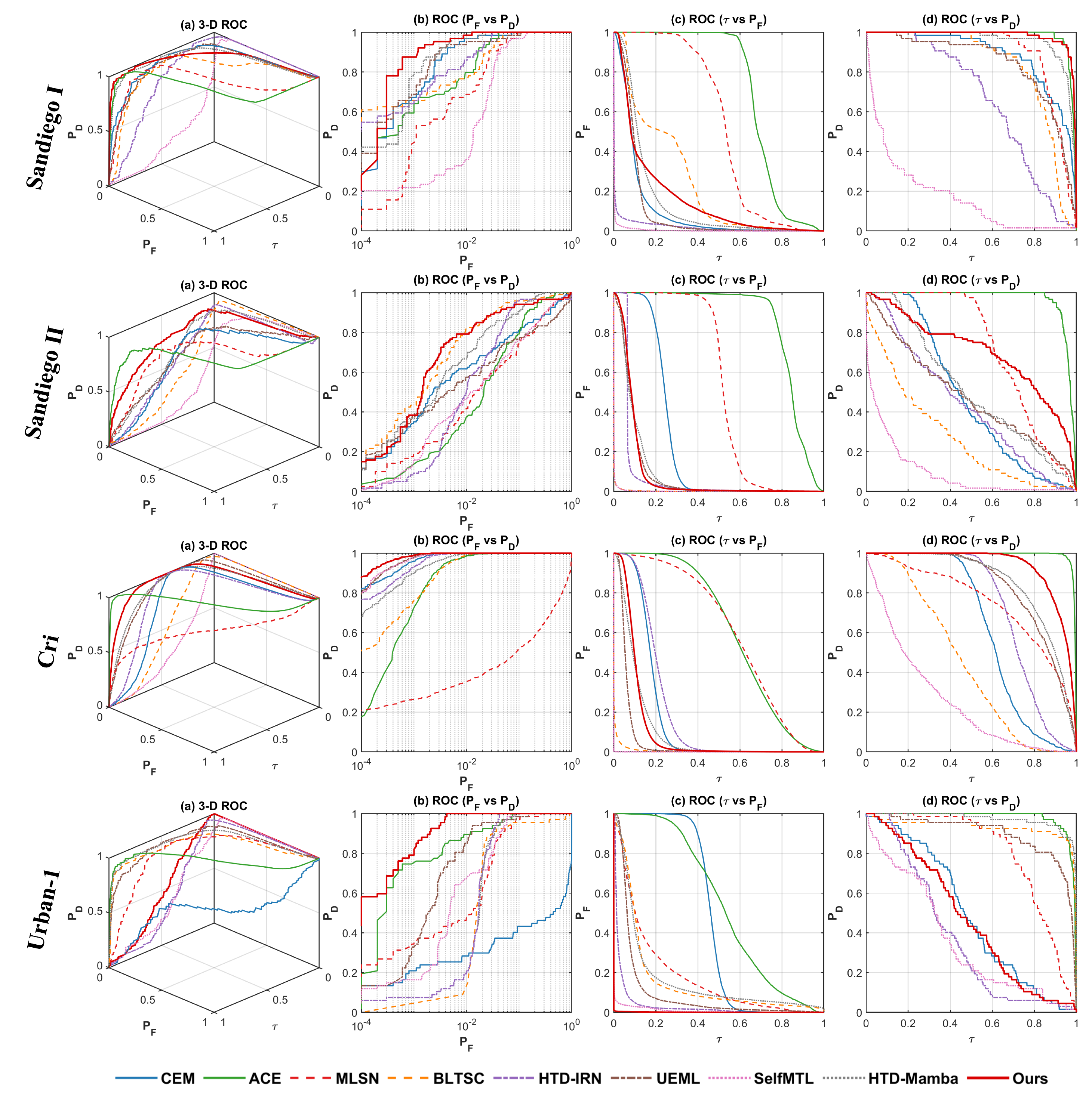}
	\caption{ROC curves for each detection method on the four datasets, from left to right: 3-D ROC curves, 2-D ROC curves $(P_F,P_D)$, 2-D ROC curves $(\tau,P_F)$, and 2-D ROC curves $(\tau,P_D)$.}
	\label{fig:roc_curves}
\end{figure*}

\begin{figure*}[!t]
	\centering
	\includegraphics[width=0.9\linewidth]{\myfigpath/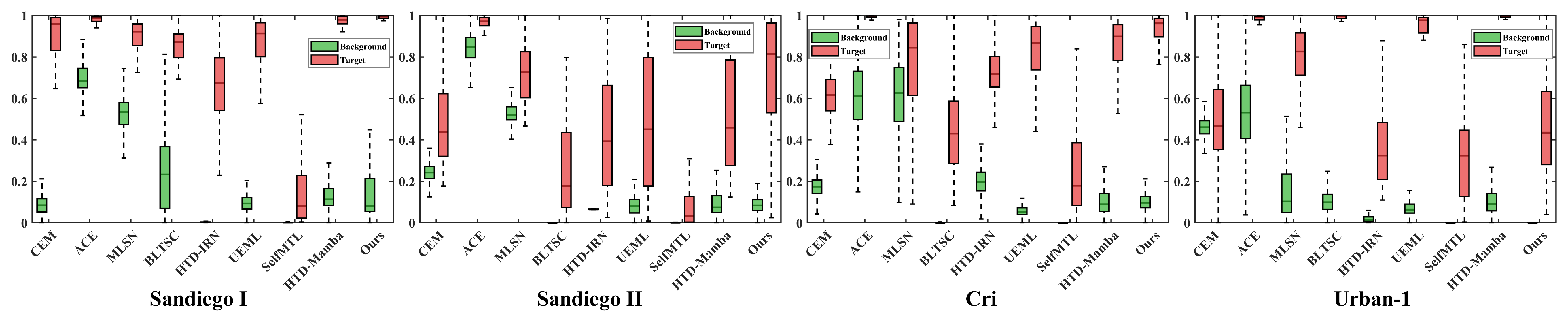}
	\caption{Target--background separation diagrams of competing methods on four datasets.}
	\label{fig:boxplots}
\end{figure*}

\subsubsection{Evaluation Metrics}
We adopt threshold-invariant ROC-based metrics, where $P_d$, $P_f$, and $\tau$ denote the detection probability, false alarm rate, and threshold, respectively. Besides $AUC_{(P_f, P_d)}$, $AUC_{(\tau, P_d)}$, and $AUC_{(\tau, P_f)}$, two additional metrics are used:
\begin{equation}
	AUC_{OA} = AUC_{(P_f, P_d)} + AUC_{(\tau, P_d)} - AUC_{(\tau, P_f)},
\end{equation}
\begin{equation}
	AUC_{SNPR} = \frac{AUC_{(\tau, P_d)}}{AUC_{(\tau, P_f)}}.
\end{equation}
Here, $AUC_{OA}$ measures the overall confidence margin between target and background, while $AUC_{SNPR}$ reflects the signal-to-noise probability ratio. ROC curves and target-background boxplots are further provided for visual comparison.

\subsubsection{Comparison Methods}
We compare SpecMamba with eight representative methods, including classical detectors CEM \cite{Du2003} and ACE \cite{Kraut1999}, deep learning and meta-learning methods MLSN \cite{wang2022ml}, BLTSC \cite{xie2020bl}, and HTD-IRN \cite{shen2023ht}, as well as recent advanced models UEML \cite{LIU2026112030}, SelfMTL \cite{SelfMTL}, and HTD-Mamba \cite{HTDMamba}. For fairness, all competing methods are implemented with the parameter settings recommended in their original papers.

\subsubsection{Implementation Details}
The proposed SpecMamba is implemented in PyTorch and trained on a single NVIDIA RTX 2080Ti GPU. For preprocessing, hyperspectral cubes are divided into local spatial-spectral patches with a spatial window size of $5 \times 5$. In the \textbf{DCTMA} module, the frequency partition ratios are set to $\rho_L=0.25$ and $\rho_M=0.60$.

During meta-training on the source domain, a 10-way 2-shot episodic strategy is adopted. The backbone is trained for 10,000 iterations using AdamW with a learning rate of $1 \times 10^{-4}$, weight decay of $1 \times 10^{-2}$, and a batch size of 32 episodes.
For test-time adaptation on the target domain, the model is optimized for 50 iterations. In the \textbf{SSPLM} module, the uncertainty-aware thresholds are set to the 0.95 quantile for targets ($\tau_{pos}$) and the 0.05 quantile for background ($\tau_{neg}$). The loss weights are set to $\beta=1.0$, $\gamma=0.1$, and $\eta=0.4$, while the physical prior rectification coefficient $\lambda$ is fixed at 0.7.

\subsection{Main Results and Analysis}
\label{subsec:results}

The quantitative results on the four benchmark datasets are reported in Table~\ref{tab:quantitative}. Under the 10-way 2-shot cross-domain setting, SpecMamba achieves the best overall performance, especially in terms of $AUC_{(P_f, P_d)}$, $AUC_{(\tau, P_d)}$, and $AUC_{OA}$. In particular, it obtains the highest $AUC_{(P_f, P_d)}$ on all four datasets, demonstrating strong robustness in detecting unseen target materials under limited supervision. On Urban-1, SpecMamba also shows superior background suppression, achieving an $AUC_{SNPR}$ of \textbf{464.73765} and a low $AUC_{(\tau, P_f)}$ of \textbf{0.00101}.

On San Diego I, San Diego II, and Cri, SpecMamba does not always achieve the lowest $AUC_{(\tau, P_f)}$ or the highest $AUC_{SNPR}$ compared with sparsity-driven methods such as SelfMTL and BLTSC. This reflects the trade-off between target enhancement and background suppression. While these methods suppress background responses more aggressively, they may also weaken sub-pixel targets, as indicated by their lower $AUC_{(\tau, P_d)}$. In contrast, SpecMamba preserves target-related spectral cues through PGTE, leading to higher detection confidence and consistently stronger $AUC_{OA}$, which indicates a better overall target-background margin.

The visual results in Fig.~\ref{fig:visual_maps} are consistent with the quantitative analysis. Compared with HTD-IRN and MLSN, SpecMamba produces more spatially continuous detection maps with clearer target boundaries. In addition, the DCTMA module suppresses high-frequency noise and reduces the salt-and-pepper artifacts commonly observed in classical detectors such as CEM and ACE. These results verify the effectiveness of the proposed spectral-spatial modeling and cross-gated fusion strategy.

Further evidence is provided by the ROC curves in Fig.~\ref{fig:roc_curves}. The $P_d$--$P_f$ curves of SpecMamba are consistently closer to the top-left corner, indicating higher detection probability at low false alarm rates. Meanwhile, its $\tau$--$P_f$ curves drop more rapidly and its $\tau$--$P_d$ curves remain more stable over a wide threshold range, demonstrating better background suppression and threshold robustness.

Finally, the boxplots in Fig.~\ref{fig:boxplots} show that SpecMamba maintains compact target-score distributions with high medians and clear separation from the background. Although the background range is slightly higher than that of some sparsity-driven methods in several scenes, the larger gap between target and background distributions indicates more reliable decision boundaries for complex real-world environments.

\section{Conclusion}\label{sec:intro5}

In this article, we have presented SpecMamba, a physics-guided meta-learning framework that addresses the challenge of few-shot hyperspectral target detection through the lens of parameter-efficient adaptation. Unlike conventional methods that rely on black-box fine-tuning, our approach freezes the heavy semantic backbone and introduces a lightweight Discrete Cosine Transform Mamba Adapter (DCTMA). This design achieves linear-complexity modeling of global spectral dependencies by re-conditioning spectral signatures in the frequency domain. To bridge the domain gap, we designed a Prior-Guided Tri-Encoder (PGTE) with a unique bifurcated pathway, allowing physical spectral priors to explicitly steer the alignment of the adapter while preserving the stability of the meta-learned representations. Furthermore, the proposed Self-Supervised Pseudo-Label Mapping (SSPLM) strategy facilitates robust test-time inference by refining decision boundaries via uncertainty-aware sampling and consistency regularization. Extensive experiments on multiple public datasets demonstrate that SpecMamba significantly outperforms state-of-the-art competitors in both detection accuracy and cross-domain transferability. In future work, we plan to explore domain randomization techniques to further enhance model robustness against open-set targets and extreme atmospheric variabilities in unconstrained environments.

\bibliographystyle{IEEEtran}
\bibliography{mybib.bib}

\appendix
This supplementary material provides additional experimental evidence to complement the main paper. In particular, we further analyze the proposed \textbf{SpecMamba} from two aspects: 1) parameter sensitivity under the 10-way 2-shot protocol, including contextual, adaptation, and optimization-related hyperparameters; and 2) ablation studies on the core modules and internal design choices. These results provide a more complete understanding of the robustness and effectiveness of the proposed framework, and further support the conclusions drawn in the main text.

Unless otherwise specified, all experimental settings follow those in the main paper.

\section{Additional Parameter Sensitivity Analysis}
\label{sec:supp_para_analysis}

To further evaluate the robustness of SpecMamba, we conduct a systematic sensitivity analysis on key hyperparameters involved in physical guidance, frequency decomposition, and test-time adaptation.

\subsection{Sensitivity of Contextual and Adaptation Parameters}
\label{subsec:supp_contextual}

\begin{figure}[htbp]
	\centering
	\includegraphics[width=1\linewidth]{\myfigpath/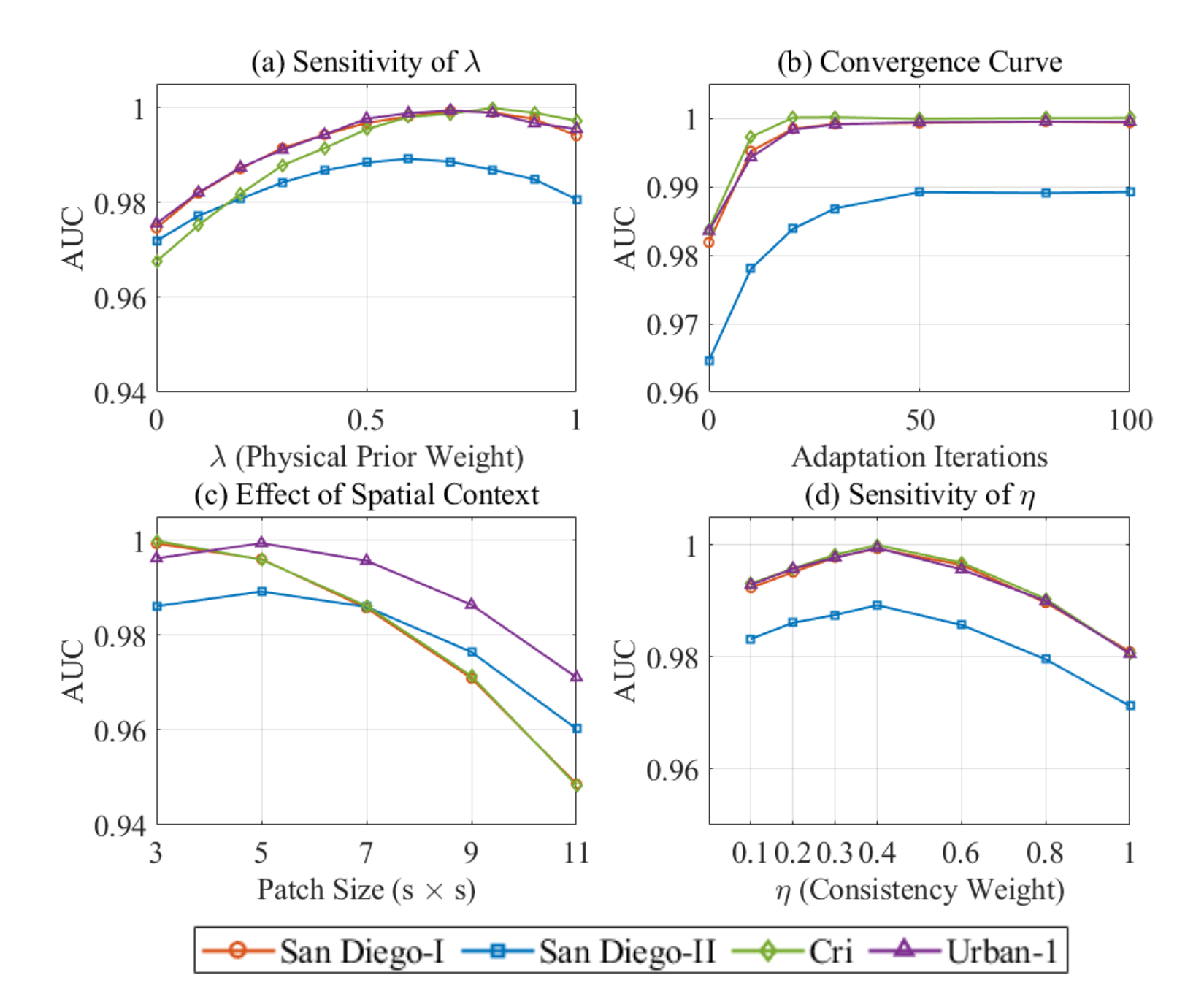}
	\caption{Fig. S1. Sensitivity analysis of SpecMamba with respect to four key parameters on the four benchmark datasets.}
	\label{fig:supp_canshu}
\end{figure}

Fig.~\ref{fig:supp_canshu} illustrates the sensitivity of four important parameters that govern physical prior guidance and test-time adaptation.

\begin{itemize}
	\item \textbf{Physical Prior Weight ($\lambda$):} As shown in Fig.~\ref{fig:supp_canshu}(a), the performance reaches its optimum when $\lambda \approx 0.7$, indicating that laboratory spectral priors effectively alleviate prototype drift in the 2-shot setting.
	
	\item \textbf{Adaptation Iterations:} Fig.~\ref{fig:supp_canshu}(b) shows that SpecMamba converges rapidly within about 50 iterations. On the challenging \textit{San Diego-II} dataset, the performance stabilizes around 0.989, demonstrating the efficiency of the lightweight adaptation strategy.
	
	\item \textbf{Spatial Context ($s \times s$):} As shown in Fig.~\ref{fig:supp_canshu}(c), a $5 \times 5$ patch achieves the best trade-off between local target localization and contextual modeling. Larger windows introduce more heterogeneous background information and slightly degrade detection performance.
	
	\item \textbf{Consistency Weight ($\eta$):} Fig.~\ref{fig:supp_canshu}(d) shows a clear convex trend, with the best performance achieved at $\eta = 0.4$. This suggests that a moderate consistency constraint is beneficial for balancing pseudo-label supervision and self-supervised regularization.
\end{itemize}

\subsection{Analysis of Frequency Partition Ratios}
\label{subsec:supp_freq_ratio}

The performance of the proposed Discrete Cosine Transform Mamba Adapter (DCTMA) depends on how spectral coefficients are partitioned into low-, mid-, and high-frequency components. To study this effect, we evaluate different combinations of the low-frequency boundary $\rho_L$ and the mid-frequency boundary $\rho_M$, as reported in Table~\ref{tab:supp_freq_ratio}.

The results show that the best performance is consistently obtained at $\rho_L = 0.25$ and $\rho_M = 0.60$ across all datasets. For example, the corresponding AUC values on \textit{AV-I} and \textit{Urban-1} reach \textbf{0.9936} and \textbf{0.9994}, respectively. This observation is consistent with the physical energy distribution of hyperspectral signatures: low-frequency coefficients mainly describe smooth background continua, while discriminative target-related absorption patterns are concentrated in the mid-frequency range. In contrast, using a larger low-frequency range (e.g., $\rho_L = 0.35$) suppresses more informative fine-grained components and leads to a slight performance drop.

\begin{table}[htbp]
	\centering
	\footnotesize
	\caption{Table S1. Sensitivity analysis of frequency partition ratios ($\rho_L, \rho_M$) in the DCTMA module (AUC).}
	\label{tab:supp_freq_ratio}
	\renewcommand\arraystretch{1.3}
	\setlength{\tabcolsep}{6.5pt}
	\begin{tabular}{cc|cccc}
		\toprule[1.25pt]
		$\rho_L$ & $\rho_M$ & \textbf{AV-I} & \textbf{AV-II} & \textbf{Cri} & \textbf{Urban-1} \\ \midrule
		0.20 & 0.50 & 0.9894 & 0.9821 & 0.9968 & 0.9912 \\
		\textbf{0.25} & \textbf{0.60} & \textbf{0.9936} & \textbf{0.9891} & \textbf{0.9997} & \textbf{0.9994} \\
		0.30 & 0.65 & 0.9921 & 0.9862 & 0.9989 & 0.9982 \\
		0.35 & 0.70 & 0.9897 & 0.9815 & 0.9961 & 0.9954 \\ \bottomrule[1.25pt]
	\end{tabular}
\end{table}

\subsection{Joint Sensitivity of Thresholds and Loss Weights}
\label{subsec:supp_joint}

\begin{figure}[htbp]
	\centering
	\includegraphics[width=1\linewidth]{\myfigpath/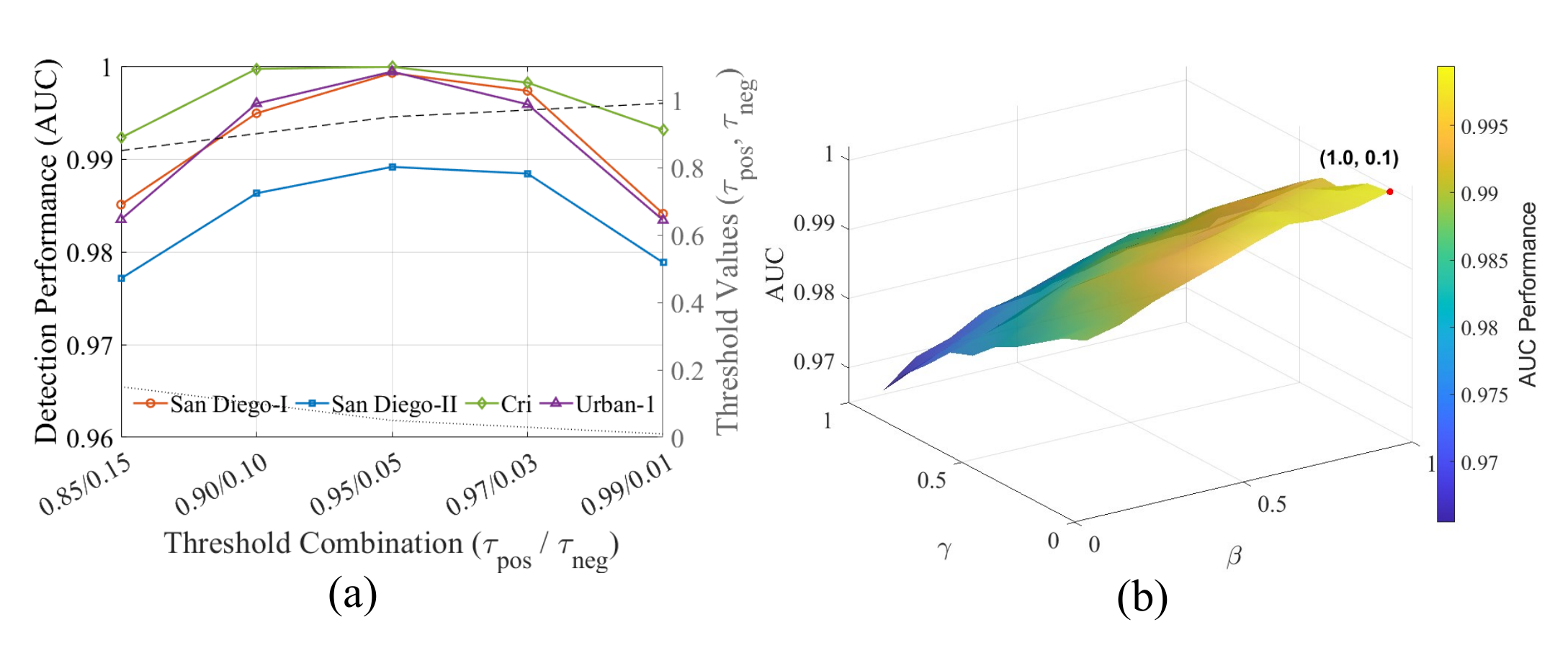}
	\caption{Fig. S2. Sensitivity analysis of pseudo-label confidence thresholds and joint loss weights in SpecMamba.}
	\label{fig:supp_canshu2}
\end{figure}

We further analyze the joint effects of pseudo-label confidence thresholds and optimization weights, as shown in Fig.~\ref{fig:supp_canshu2}.

Fig.~\ref{fig:supp_canshu2}(a) presents the sensitivity of the positive and negative confidence thresholds, $\tau_{pos}$ and $\tau_{neg}$. The combination \textbf{0.95/0.05} yields the best overall performance, indicating that this setting provides an appropriate uncertainty margin for reliable pseudo-label selection. Among the evaluated datasets, \textit{Cri} shows the lowest sensitivity to threshold variation, with its AUC remaining close to 1.0 even under less strict settings, which reflects the high spectral purity and separability of its targets.

Fig.~\ref{fig:supp_canshu2}(b) reports the joint sensitivity of the classification loss weight $\beta$ and the physical consistency weight $\gamma$ through a $10 \times 10$ grid search. The performance forms a stable high-response region centered around $(\beta=1.0, \gamma=0.1)$, where the peak AUC on \textit{Urban-1} reaches 0.9994. The rapid degradation as $\beta \rightarrow 0$ indicates that semantic discrimination remains the primary driving factor for detection, whereas the relatively smooth variation along $\gamma$ suggests that the physical prior acts as a robust corrective constraint over a broad range.

\section{Additional Ablation Studies}
\label{sec:supp_ablation}

To further validate the necessity and internal design of the proposed \textbf{SpecMamba}, we provide additional ablation studies in this supplementary material. Specifically, we first examine the contributions of the three core modules, namely \textbf{DCTMA}, \textbf{PGTE}, and \textbf{SSPLM}, and then analyze several fine-grained design choices within DCTMA and SSPLM. For brevity, \textit{San Diego I} and \textit{San Diego II} are denoted as \textit{AV-I} and \textit{AV-II}, respectively.

\subsection{Effectiveness of Core Modules}
\label{subsec:supp_ablation_main}

We evaluate four variants under the 10-way 2-shot protocol to isolate the contributions of the three key modules. The results are reported in Table~\ref{tab:supp_ablation_main}.

\begin{table}[!htbp]
	\centering
	\footnotesize
	\caption{Table S2. Ablation analysis of the core components (AUC) across four benchmark datasets.}
	\label{tab:supp_ablation_main}
	\renewcommand\arraystretch{1.2}
	\setlength{\tabcolsep}{3pt}
	\begin{tabular}{l|ccc|cccc}
		\toprule[1.25pt]
		\textbf{Variant} & \textbf{DCTMA} & \textbf{PGTE} & \textbf{SSPLM} & \textbf{AV-I} & \textbf{AV-II} & \textbf{Cri} & \textbf{Urban-1} \\ \midrule
		Case 1 & \checkmark & & & 0.9886 & 0.9673 & 0.9876 & 0.9876 \\
		Case 2 & & \checkmark & & 0.9936 & 0.9718 & 0.9877 & 0.9682 \\
		Case 3 & \checkmark & \checkmark & & 0.9864 & 0.9731 & 0.9886 & 0.9832 \\
		\textbf{Ours} & \checkmark & \checkmark & \checkmark & \textbf{0.9992} & \textbf{0.9891} & \textbf{0.9999} & \textbf{0.9994} \\
		\bottomrule[1.25pt]
	\end{tabular}
\end{table}

The results in Table~\ref{tab:supp_ablation_main} show that each module contributes to the final performance.

\textbf{DCTMA.} Case 1 demonstrates that the frequency-aware spectral-spatial adapter already provides a strong baseline by suppressing noise and improving spectral representation quality.

\textbf{PGTE.} Compared with Case 1, Case 2 confirms that the laboratory prior serves as an effective anchor in the 2-shot setting, alleviating prototype drift caused by noisy support samples. When DCTMA and PGTE are combined (Case 3), the model becomes more stable overall, although slight optimization competition may still occur on certain data distributions such as AV-I.

\textbf{SSPLM.} The full model achieves the best results on all datasets, indicating that rapid test-time adaptation is essential for aligning the detector to the target-background distribution of each query scene and refining the decision boundary under domain shift.

\subsection{Analysis of Module Design Details}
\label{subsec:supp_ablation_sub}

We further investigate the internal design choices of DCTMA and SSPLM. The corresponding results are listed in Table~\ref{tab:supp_ablation_sub}.

\begin{table}[!htbp]
	\centering
	\footnotesize
	\caption{Table S3. Analysis of module design details (AUC).}
	\label{tab:supp_ablation_sub}
	\renewcommand\arraystretch{1}
	\setlength{\tabcolsep}{8pt}
	\begin{tabular}{l|l|cc}
		\toprule[1.25pt]
		\textbf{Module} & \textbf{Sub-Strategy} & \textbf{AV-II} & \textbf{Urban-1} \\ \midrule
		\multirow{2}{*}{DCTMA} & w/o Freq. Masking & 0.9745 & 0.9812 \\
		& \textbf{Multi-Freq Partition} & \textbf{0.9891} & \textbf{0.9994} \\ \midrule
		\multirow{3}{*}{SSPLM} & Only Spatial Aug. & 0.9812 & 0.9904 \\
		& Only Spectral Aug. & 0.9834 & 0.9921 \\
		& \textbf{Hybrid Aug.} & \textbf{0.9891} & \textbf{0.9994} \\
		\bottomrule[1.25pt]
	\end{tabular}
\end{table}

\textbf{Impact of frequency masking in DCTMA.} Removing explicit frequency masking leads to performance drops on both AV-II and Urban-1. This verifies that high-frequency coefficients contain non-negligible sensor noise, which can interfere with material-specific discrimination if not properly separated. The proposed multi-frequency partition better preserves reliable mid-frequency absorption information for detection.

\textbf{Effectiveness of hybrid augmentation in SSPLM.} Using only spatial or only spectral augmentation degrades the performance compared with the full hybrid strategy. Spatial augmentation improves robustness to geometric disturbance, while spectral perturbation reduces sensitivity to non-discriminative intensity variation. Their combination provides stronger regularization during test-time adaptation and results in the best detection accuracy.

\end{document}